# Beetle Swarm Optimization Algorithm:Theory and Application

Tiantian Wang[a], Long Yang[b]

[a]Engineering College,Ocean University of China,238Songling Road ,Laoshan District,Qingdao 266100, Shandong,China

[b]Chemistry Experimental Center, Xihua University,9999Hongguang Road, Pidu District,Chengdu 610039,Sichuan,China

**Abstract.** In this paper, a new meta-heuristic algorithm, called beetle swarm optimization (BSO) algorithm, is proposed by enhancing the performance of swarm optimization through beetle foraging principles. The performance of 23 benchmark functions is tested and compared with widely used algorithms, including particle swarm optimization (PSO) algorithm, genetic algorithm (GA) and grasshopper optimization algorithm (GOA). Numerical experiments show that the BSO algorithm outperforms its counterparts. Besides, to demonstrate the practical impact of the proposed algorithm, two classic engineering design problems, namely, pressure vessel design problem and himmelblau's optimization problem, are also considered and the proposed BSO algorithm is shown to be competitive in those applications.

## 1. Introduction

In the past decade, various optimization algorithms have been proposed and applied to different research fields. Procedures may vary to solve different optimization problems, but the following questions need to be considered in advance before selecting the optimization algorithm: (1) Parameters of the problem. The problem can be divided into continuous or discrete depending on the parameters. (2) Constraints of variables. Optimization problems can be classified into constrained and unconstrained ones based on the type of constraints[1]. (3) The cost function of a given problem. The problem can be divided into single-objective and multi-objective problems[2]. Based on the above three points, we need to select the optimization algorithm according to the parameter type, constraint and target number.

The development of optimization algorithms is relatively mature at present, and many excellent optimization algorithms have been applied in various fields. We can divide the optimization algorithms into two categories: gradient-based methods and meta-heuristic algorithms. For simple problems such as continuous and linear problems, some classical algorithm gradient algorithms can be utilized, such as Newton's method[3], truncated gradient method[4], gradient descent method[5],etc. For more complex problems, meta-heuristics such as genetic algorithm[6], ant colony algorithm[7]and particle swarm optimization algorithm[8]can be considered. And the meta heuristic algorithm becomes very popular because of its stability and flexibility and its ability to better avoid local optimization[9].

People usually divide the meta-heuristic algorithm into three types, which are based on the principles of biological evolution, population and physical phenomena. The evolutionary approach is inspired by the concept of natural evolution. The population based optimization algorithm is mainly inspired by the social behavior of animal groups, while the physical phenomenon based method mainly imitates the physical rules of the universe. Table 1 summarizes the algorithms included in each category.

**Table 1** Algorithm Classification

| | | |
|---|---|---|
| Meta-heuristic Algorithms | Evolutionary Algorithms | Genetic Algorithm[6] |
| | | Evolution Strategies[11] |
| | | Probability-Based Incremental Learning[12] |
| | | Genetic Programming[13] |
| | | Biogeography-Based Optimizer[14] |
| | Physics-based Algorithms | Simulated Annealing[15] |
| | | Gravitational Local Search[16] |
| | | Big-Bang Big-Crunch[17] |
| | | Gravitational Search Algorithm[18] |
| | | Charged System Search[19] |
| | | Central Force Optimization[20] |
| | | Artificial Chemical Reaction Optimization Algorithm[21] |
| | | Black Hole algorithm[22] |
| | | Ray Optimization algorithm[23] |
| | | Small-World Optimization Algorithm[24] |
| | | Galaxy-based Search Algorithm[25] |
| | | Curved Space Optimization[26] |
| | Swarm-based Algorithms | particle swarm optimization algorithm[8] |
| | | Honey Bees Optimization Algorithm[27] |
| | | Artificial Fish-Swarm Algorithm[28] |
| | | Termite Algorithm[29] |
| | | Wasp Swarm Algorithm[30] |
| | | Monkey Search[31] |
| | | Bee Collecting Pollen Algorithm[32] |
| | | Cuckoo Search[33] |
| | | Dolphin Partner Optimization[34] |
| | | Firefly Algorithm[35] |
| | | Bird Mating Optimizer[36] |
| | | Fruit fly Optimization Algorithm[37] |

In face of so many existing meta-optimization algorithms, a concern naturally rises. So far, there have been many different types of optimization algorithms. Why do we need more algorithms? We will mention that there is no free lunch (NFL)[38] theorem, no matter how smart or how clumsy the optimization algorithm is, their performance is logically equivalent. That is, there is no optimization algorithm that can solve all optimization problems. This theorem makes the number of algorithms increase rapidly over the past decade, which is one of the motivations of this paper.

In this paper, a new optimization, namely Beetle Swarm Optimization (BSO) algorithm, is proposed by combining beetle foraging mechanism with group optimization algorithm. The rest of the paper is structured as follows. Section 2 describes the Beetle Swarm Optimization algorithm developed in this study. Section 3 tests the performance of the algorithm on the unimodal functions, multimodal functions and fixed-dimension multimodal functions. Section 4 applies the BSO algorithm to the multi-objective problems to further test the performance of the algorithm. Section 5 draws conclusions.

# 2 Beetle Swarm Optimization(BSO)

## 2.1 Beetle Antennae Search Algorithm

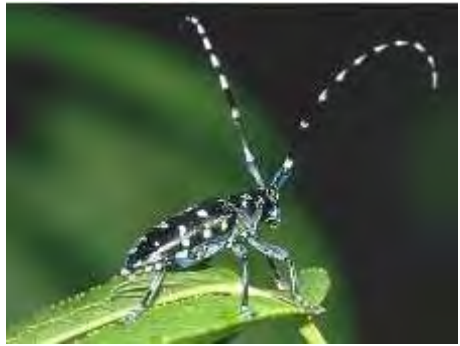

**Fig.1** longhorn beetle

Insects have a shifting chemical sensory system that senses various environmental stimuli and guides their behavior[39,40],The antennae of insects are important chemical receptors. They mainly play olfactory and tactile effects, and some even have an auditory function. They can help insects communicate, find the opposite sex, find food and choose spawning sites[41].People often use this property of insects to release substances with specific volatile odors to attract or evade insects harmful to plants[42].The long-horned beetle shown in fig.1 is characterized by extremely long antennae, sometimes up to four times the length of its body. This kind of long antennae has two basic functions: one is to explore the surrounding environment. For example, when encountering an obstacle, the feeler can perceive its size, shape and hardness. The second is to capture the smell of food or find potential mates by swinging the body's antenna. When a higher concentration of odor is detected on one side of the antenna, the beetle will rotate in the same direction, otherwise it will turn to the other side. According to this simple principle, beetles can effectively find food[43].A meta-heuristic optimization algorithm based on the search behavior of long-horned beetles was proposed by Jiang X et al. [43,44].Similar to genetic algorithms, particle swarm algorithms, etc., Beetle Antennae Search (BAS)Algorithm can automatically realize the optimization process without knowing the specific form of the function and gradient information. The major advantage of the BAS is the lesser complexity involved in its design and in its ability to solve the optimization problem in less time since its individual number is only one.

When using BAS to optimize nonlinear systems, a simple two-step building procedure is employed as follows: (i) model the searching behavior; (ii) formulate the behavior of detecting. In this section, the position of beetle at time t (t=1,2,...) is denoted as $x^t$, denote the concentration of odor at position $x$ to be $f(x)$ known as a fitness function, where the maximum (or minimum) value corresponds to the point of odor source.

Mathematically, BAS model is stated as follows. The random search directions of beetles are shown as follows[43]:

$$\vec{b} = \frac{rands(n,1)}{\|rands(n,1)\|} \tag{1}$$

where $rands(.)$ denote the random function, and $n$ indicates the space dimension. Then create the beetle's left and right spatial coordinates[43,45]:

$$\begin{aligned} x_{rt} &= x^t + d_0 * \vec{b}/2 \\ x_{lt} &= x^t - d_0 * \vec{b}/2 \end{aligned} \tag{2}$$

where $x_{rt}$ represents the position coordinates of the right antennae at time t ,and $x_{lt}$ represents the coordinates of the left antennae at time t. $d_0$ represents the distance between two antennae. Use the fitness function value to represent the scent intensity at the right and left antennae, we denote them as $f(x_{rt})$ and $f(x_{lt})$.

In the next step, we set the beetle's iterative mechanism to formulate the detect behavior, the model as follows[43]:

$$x^{t+1} = x^t + \delta^t * \vec{b} * sign(f(x_{rt}) - f(x_{lt})) \tag{3}$$

where $\delta^t$ represents the step factor, the step size usually decreases as the number of iterations increases. $sign(.)$ represents a sign function.

It is worth pointing out that searching distance $d_0$ and $\delta$. In general, we set the initial step length as a constant, and the initial step length increases as the fitness function dimension increases. To simplify the parameter turning further more, we also construct the relationship between searching distance $d$ and step size $\delta$ as follows[44]:

$$\delta^t = c_1 \delta^{t-1} + \delta^0 \quad \text{or} \quad \delta^t = eta * \delta^{t-1} \tag{4}$$
$$d^t = \delta^t / c_2 \tag{5}$$

where $c_1, c_2$ and $eta$ are constants to be adjusted by designers, we recommend eta's value is 0.95. Fig. 2 shows a schematic diagram of the movement of the beetle, which can help understand the optimization process.

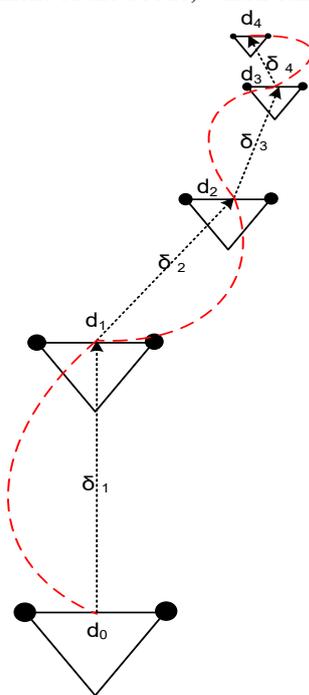

**Fig.2** Beetle's four-step optimization process. The black triangle represents the beetle, the black solid circles on both sides represent the beard of the beetle, $d_i$ $(i=1,2,3,4)$ represents the distance between the two antennaes, $\delta_i$ represents the step length, and the red dashed line represents the trajectory of the fitness function.

## 2.2 Beetle Swarm Optimization Algorithm

With the continuous deepening of the experiment, we found that the performance of the BAS algorithm in dealing with high-dimensional functions is not very satisfactory, and the iterative result is very dependent on the initial position of the beetle. In other words, the choice of initial position greatly affects the efficiency and effectiveness of optimization. Inspired by the swarm optimization algorithm, we have made further improvements to the BAS algorithm by expanding an individual to a group. That is the beetle swarm optimization (BSO) algorithm we will introduce.

In this algorithm, each beetle represents a potential solution to the optimization problem, and each beetle corresponds to an fitness value determined by the fitness function. Similar to the particle swarm algorithm, the beetles also share information, but the distance and direction of the beetles are determined by their speed and the intensity of the information to be detected by their long antennae.

In mathematical form, we borrowed the idea of particle swarm algorithm. There is a population of n beetles represented as $X = (X_1, X_2, \cdots, X_n)$ in an S-dimensional search space, where the $i$ th beetle represents an S-dimensional vector $X_i = (x_{i1}, x_{i2}, \cdots, x_{iS})^T$, represents the position of the $i$ th beetle in the S-dimensional search space, and also represents a potential solution to the problem. According to the target function, the fitness value of each beetle position can be calculated. The speed of the $i$ th beetle is expressed as $V_i = (V_{i1}, V_{i2}, \cdots, V_{iS})^T$. The individual extremity of the beetle is represented as $P_i = (P_{i1}, P_{i2}, \cdots, P_{iS})^T$, and the group extreme value of the population is represented as $P_g = (P_{g1}, P_{g2}, \cdots, P_{gS})^T$ [46]. The mathematical model for simulating its behavior is as follows:

$$X_{is}^{k+1} = X_{is}^k + \lambda V_{is}^k + (1-\lambda)\xi_{is}^k \tag{6}$$

where $s = 1, 2, \cdots, S$; $i = 1, 2, \cdots, n$; $k$ is the current number of iterations. $V_{is}$ is expressed as the speed of beetles, and $\xi_{is}$ represents the increase in beetle position movement. $\lambda$ is a positive constants.

Then the speed formula is written as[8,47,48]:

$$V_{is}^{k+1} = \omega V_{is}^k + c_1 r_1 (P_{is}^k - X_{is}^k) + c_2 r_2 (P_{gs}^k - X_{gs}^k) \tag{7}$$

where $c_1$ and $c_2$ are two positive constants, and $r_1$ and $r_2$ are two random functions in the range[0,1]. $\omega$ is the inertia weight. In the standard PSO algorithm, $\omega$ is a fixed constant, but with the gradual improvement of the algorithm, many scholars have proposed a changing inertia factor strategy[46,49,50].

This paper adopts the strategy of decreasing inertia weight, and the formula is as follows[46]:

$$\omega = \omega_{max} - \frac{\omega_{max} - \omega_{min}}{K} * k \tag{8}$$

Where $\omega_{min}$ and $\omega_{max}$ respectively represent the minimum and maximum value of $\omega$. $k$ and $K$ are the current number of iterations and the maximum number of iterations. In this paper, the maximum value of $\omega$ is set to 0.9, and the minimum value is set to 0.4[51], so that the algorithm can search a larger range at the beginning of evolution and find the optimal solution area as quickly as possible. As $\omega$ gradually decreases, the beetle's speed decreases and then enters local search.

The $\xi$ function, which defines the incremental function, is calculated as follows:

$$\xi_{is}^{k+1} = \delta^k * V_{is}^k * sign(f(X_{rs}^k) - f(X_{ls}^k)) \tag{9}$$

In this step, we extend the update (3) to a high dimension. $\delta$ indicates step size. The search behaviors of the right antenna and the left antenna are respectively expressed as:

$$\begin{aligned} X_{rs}^{k+1} &= X_{rs}^k + V_{is}^k * d/2 \\ X_{ls}^{k+1} &= X_{ls}^k - V_{is}^k * d/2 \end{aligned} \tag{10}$$

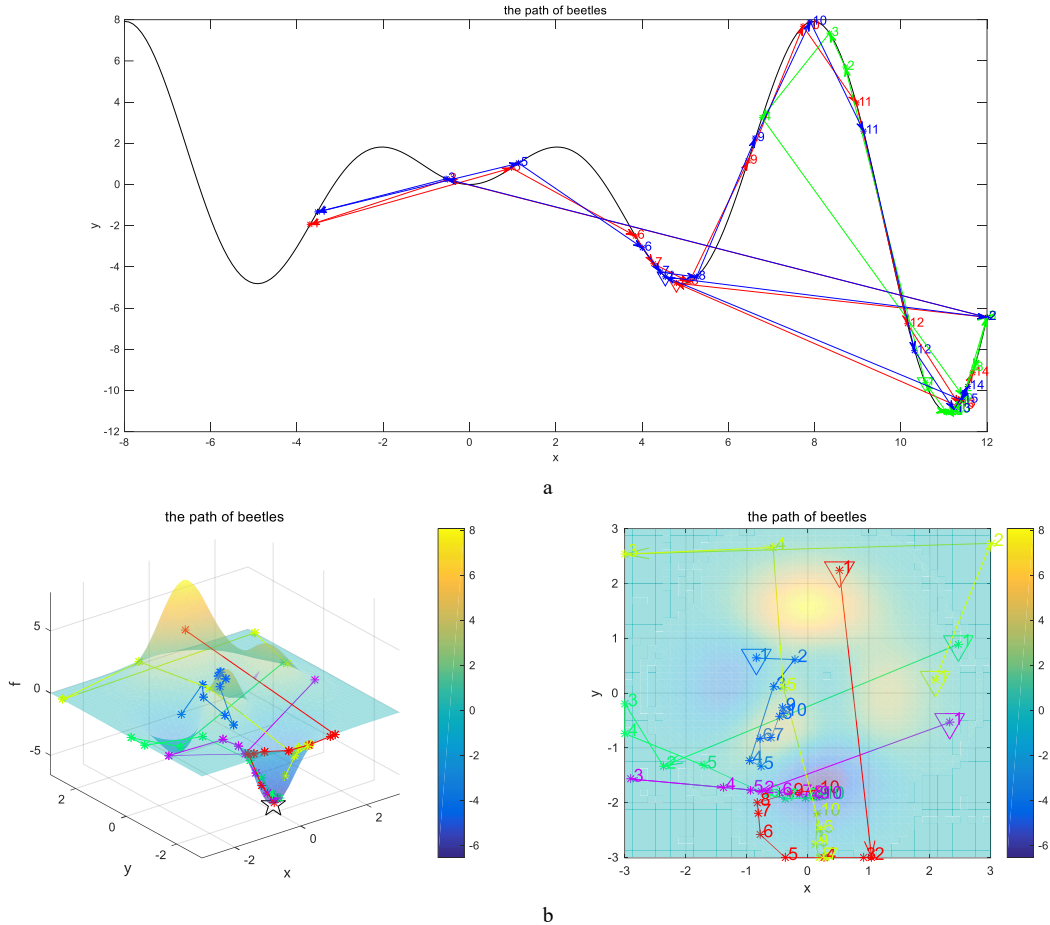

**Fig.3** Beetles Search Path in 2D Space(a) and 3D Space(b)

Fig.3 shows the trajectories of the beetle swarm in two-dimensional and three-dimensional space, respectively. To represent the search path more visually, we used a small population size and showed the location change process of 10 iterations in 3D space. Because factors such as step length and inertial weight coefficient are decreasing in the iterative process, the algorithm will not converge to the target point too quickly, thus avoiding the group falling into the local optimum greatly.

The BSO algorithm first initializes a set of random solutions. At each iteration, the search agent updates its location based on its own search mechanism and the best solution currently available. The combination of these two parts can not only accelerate the population's iteration speed, but also reduce the probability of the population falling into the local optimum, which is more stable when dealing with high-dimensional problems.

The pseudo code of the BSO algorithm is presented.

**Procedure:**
Initialize the swarm $X_i (i = 1, 2, ..., n)$
Initialize population speed $v$
Set step size $\delta$, speed boundary $v_{max}$ and $v_{min}$, population size $sizepop$ and maximum number of iterations $K$
Calculate the fitness of each search agent
**While**( $k < K$ )
  Set inertia weight $\omega$ using Eq.(8)
  Update $d$ using Eq.(5)
  **for** each search agent
    Calculate $f(X_{rs})$ and $f(X_{ls})$ using Eq.(10)
    Update the incremental function $\xi$ by the Eq.(9)
    Update the speed formula $V$ by the Eq.(7)
    Update the position of the current search agent by the Eq.(6)
  **end for**
  Calculate the fitness of each search agent $f(x)$
  Record and store the location of each search agent

```
for each search agent
    if f(x) < f_pbest
        f_pbest = f(x)
    end if
    if f(x) < f_gbest
        f_gbest = f(x)
    end if
end for
Update x* if there is a better solution
Update step factor δ by the Eq.(4)
end while
Return x_best, f_best
```

In theory, the BSO algorithm includes exploration and exploitation ability, so it belongs to global optimization. Furthermore, the linear combination of speed and beetle search enhances the rapidity and accuracy of population optimization and makes the algorithm more stable. In the next section, we will examine the performance of the proposed algorithm through a set of mathematical functions.

## 3. Results and Discussion

In the optimization field, a set of mathematical functions with optimal solutions is usually used to test the performance of different optimization algorithms quantitatively. And the test functions should be diverse so that the conclusions are not too one-sided. In this paper, three groups of test functions with different characteristics are used to benchmark the performance of the proposed algorithm which are unimodal functions, multimodal functions and fixed-dimension multimodal functions[52,53,54,55,56,57]. The specific form of the function is given in table 2-4, where *Dim* represents the dimension of the function, *Range* represents the range of independent variables, that is, the range of population, and $f_{min}$ represents the minimum value of the function.

**Table2** Description of unimodal benchmark functions

| Function | Dim | Range | $f_{min}$ |
|---|---|---|---|
| $f_1(x) = \sum_{i=1}^{n} x_i^2$ | 30 | [-100,100] | 0 |
| $f_2(x) = \sum_{i=1}^{n} |x_i| + \prod_{i=1}^{n} |x_i|$ | 30 | [-10,10] | 0 |
| $f_3(x) = \sum_{i=1}^{n} \left( \sum_{j-1}^{i} x_j \right)^2$ | 30 | [-100,100] | 0 |
| $f_4(x) = \max_i \{|x_i|, 1 \leq i \leq n\}$ | 30 | [-100,100] | 0 |
| $f_5(x) = \sum_{i=1}^{n-1} [100(x_{i+1} - x_i^2)^2 + (x_i - 1)^2]$ | 30 | [-30,30] | 0 |
| $f_6(x) = \sum_{i=1}^{n} ([x_i + 0.5])^2$ | 30 | [-100,100] | 0 |
| $f_7 = \sum_{i=1}^{n} i x^4 + random[0,1)$ | 30 | [-1.28,1.28] | 0 |

**Table3** Description of multimodal benchmark functions

| Function | Dim | Range | $f_{min}$ |
|---|---|---|---|
| $f_8(x) = \sum_{i=1}^{n} -x_i \sin(\sqrt{|x_i|})$ | 30 | [-500,500] | -418.9829*Dim |
| $f_9(x) = \sum_{i=1}^{n} [x_i^2 - 10\cos(2\pi x_i) + 10]$ | 30 | [-5.12,5.12] | 0 |
| $f_{10}(x) = -20\exp(-0.2\sqrt{\frac{1}{n}\sum_{i=1}^{n} x_i^2}) - \exp(\frac{1}{n}\sum_{i=1}^{n} \cos(2\pi x_i)) + 20 + e$ | 30 | [-32,32] | 0 |
| $f_{11}(x) = \frac{1}{4000}\sum_{i=1}^{n} x_i^2 - \prod_{i=1}^{n} \cos(\frac{x_i}{\sqrt{i}}) + 1$ | 30 | [-600,600] | 0 |

| Function | | Dim | Range | $f_{min}$ |
|---|---|---|---|---|
| $f_{12}(x) = \frac{\pi}{n}\{10\sin(\pi y_1) + \sum_{i=1}^{n-1}(y_i-1)^2[1+10\sin^2(\pi y_{i+1})] + (y_n-1)^2\}$ $+ \sum_{i=1}^{n} u(x_i,10,100,4)$ $y_i = 1 + \frac{x_i+1}{4}, u(x_i,a,k,m) = \begin{cases} k(x_i-a)^m & x_i > a \\ 0 & -a < x_i < a \\ k(-x_i-a)^m & x_i < -a \end{cases}$ | | 30 | [-50,50] | 0 |
| $f_{13}(x) = 0.1\{\sin^2(3\pi x_1) + \sum_{i=1}^{n}(x_i-1)^2[1+\sin^2(3\pi x_i+1)] + (x_n-1)^2[1+\sin^2(2\pi x_n)]\}$ $+ \sum_{i=1}^{n} u(x_i,5,100,4)$ | | 30 | [-50,50] | 0 |

**Table 4** Description of fixed-dimension multimodal benchmark functions

| Function | Dim | Range | $f_{min}$ |
|---|---|---|---|
| $f_{14}(x) = (\frac{1}{500} + \sum_{j=1}^{25}(j + \sum_{i=1}^{2}(x_i-a_{ij})^6)^{-1})^{-1}$ | 2 | [-65,65] | 0.9980 |
| $f_{15}(x) = \sum_{i=1}^{11}[a_i - \frac{x_1(b_i^2+b_ix_2)}{b_i^2+b_ix_3+x_4}]^2$ | 4 | [-5,5] | 0.00030 |
| $f_{16}(x) = 4x_1^2 - 2.1x_1^4 + \frac{1}{3}x_1^6 + x_1x_2 - 4x_2^2 + 4x_2^4$ | 2 | [-5,5] | -1.0316 |
| $f_{17}(x) = (x_2 - \frac{5.1}{4\pi^2}x_1^2 + \frac{5}{\pi}x_1 - 6)^2 + 10(1-\frac{1}{8\pi})\cos x_1 + 10$ | 2 | [-5,5] | 0.398 |
| $f_{18}(x) = [1 + (x_1+x_2+1)^2(19-14x_1+3x_1^2-14x_2+6x_1x_2+3x_2^2)]$ $\times [30 + (2x_1-3x_2)^2 \times (18-32x_1+12x_1^2+48x_2-36x_1x_2+27x_2^2)]$ | 2 | [-2,2] | 3 |
| $f_{19}(x) = -\sum_{i=1}^{4} c_i \exp(-\sum_{j=1}^{3} a_{ij}(x_j-p_{ij})^2)$ | 3 | [1,3] | -3.86 |
| $f_{20}(x) = -\sum_{i=1}^{4} c_i \exp(-\sum_{j=1}^{6} a_{ij}(x_j-p_{ij})^2)$ | 6 | [0,1] | -3.32 |
| $f_{21}(x) = -\sum_{i=1}^{5}[(X-a_i)(X-a_i)^T + c_i]^{-1}$ | 4 | [0,10] | -10.1532 |
| $f_{22}(x) = -\sum_{i=1}^{7}[(X-a_i)(X-a_i)^T + c_i]^{-1}$ | 4 | [0,10] | -10.4028 |
| $f_{23}(x) = -\sum_{i=1}^{10}[(X-a_i)(X-a_i)^T + c_i]^{-1}$ | 4 | [0,10] | -10.5363 |

Fig.4 shows the two-dimensional versions of unimodal function, multimodal function and fixed-dimension multimodal function respectively. The unimodal test function has only one global optimal solution, which is helpful to find the global optimal solution in the search space, and it can test the convergence speed and efficiency of the algorithm well. From fig.5, it can also be seen that the multimodal function and the fixed-dimension multimodal test function have multiple local optimal solutions, which can be used to test the algorithm to avoid the performance of the local optimal solution, and the fixed-dimension multimodal function compared with unimodal test function is more challenging.

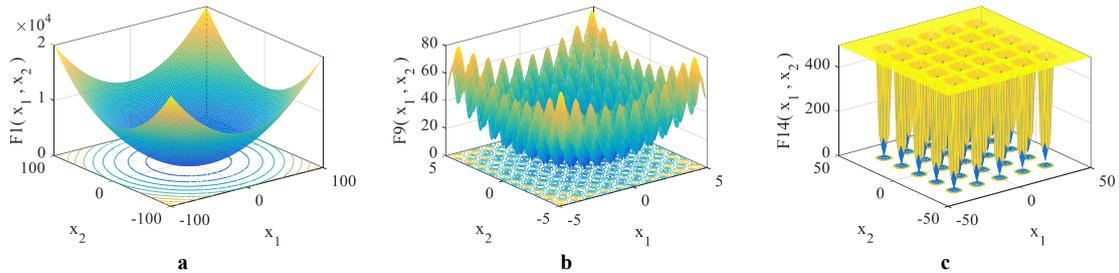

**Fig.4** 2-D version of unimodal function(a)、multimodal function(b) and fixed-dimension multimodal function(c)

In the part of qualitative analysis, six typical test functions are provided, including optimal trajectory map, contour map and convergence curve of search path. In the quantitative analysis part, 50 search agents were used, the maximum number of iterations was set to 1000, and each test function was run 30 times to generate statistical results. Quantitative evaluation was performed using the mean, standard deviation, and program performance time of three performance indicators. Statistical results are reported in Table4.BSO was compared with PSO[8],GA[6] and GOA[58].

**3.1 Qualitative Results and Discussion**

In this paper, six unimodal , multimodal and fixed-dimension multimodal functions are selected to observe the BSO algorithm's optimization behavior. In order to express the optimization trajectory more intuitively, we use five search agents.

Fig.6 shows the optimal trace of each test function, the contour map of the search path, and the convergence curves. The optimal trajectory gives the best beetle optimization route. Since the initial position of the beetle is randomly generated, the optimal trajectory may be different when reproducing the result. The contour map of the search path can more intuitively display the beetle's trajectory, and connecting the same $z$-values on the $x$, $y$ plane makes it easier to observe beetle movements. The convergence curve shows the function value of the best solution obtained so far.

From Fig.6 it can be seen that beetles gradually move to the best point and eventually gather around the global best point. This phenomenon can be observed in unimodal, multimodal, and fixed-dimension multimodal functions. The results show that the BSO algorithm has a good balance between exploration and exploitation capabilities to promote the beetle to move to the global optimum. In addition, in order to more clearly represent the trajectory of the beetle, some of the function images are processed. Such as $f_{10}$, this paper selects the opposite form and can more intuitively observe the optimal trajectory.

The BSO algorithm of the beetle self-optimization mechanism has been added, which can more intelligently avoid local optimums. During the optimization process, we found that some beetles always move quickly toward the maximum value, and then reach the maximum value and then perform normal iterations. This mechanism makes the beetle cleverly avoid the local optimum during the optimization process. For unimodal and multimodal functions, the advantage of the self-optimization mechanism is even more pronounced.

Fig.5 provides a convergence curve to further prove that this mechanism can improve the search results. The convergence curve clearly shows the descending behavior of all test functions. Observe that the BSO search agent suddenly changes during the early stage of the optimization process, and then gradually converges. According to Berg et al.[59], this behavior ensures that the algorithm quickly converges to the optimal point to reduce the iteration time.

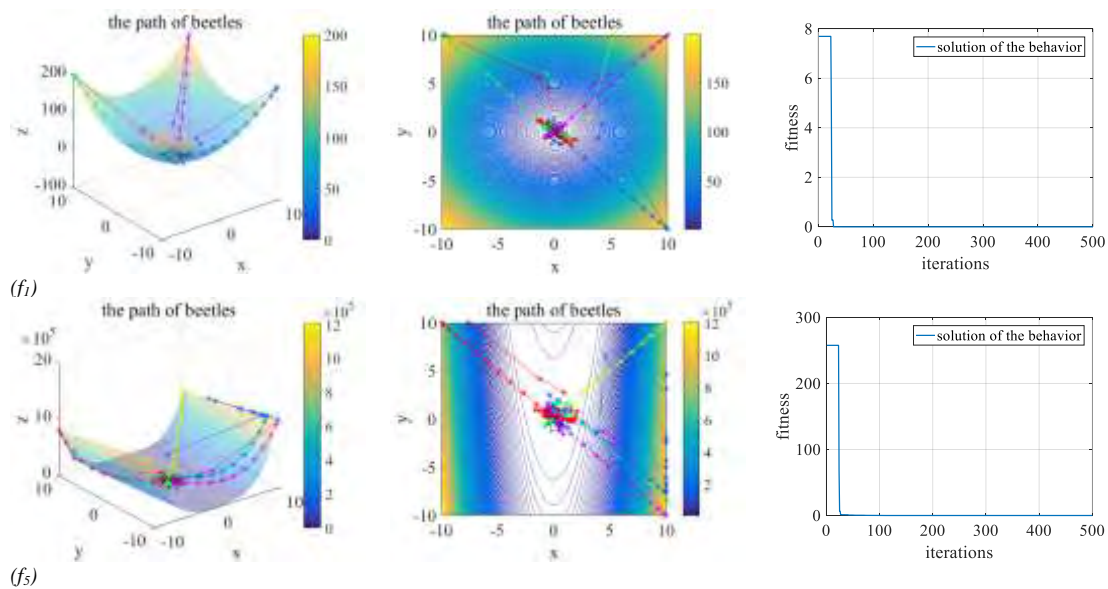

($f_1$)

($f_5$)

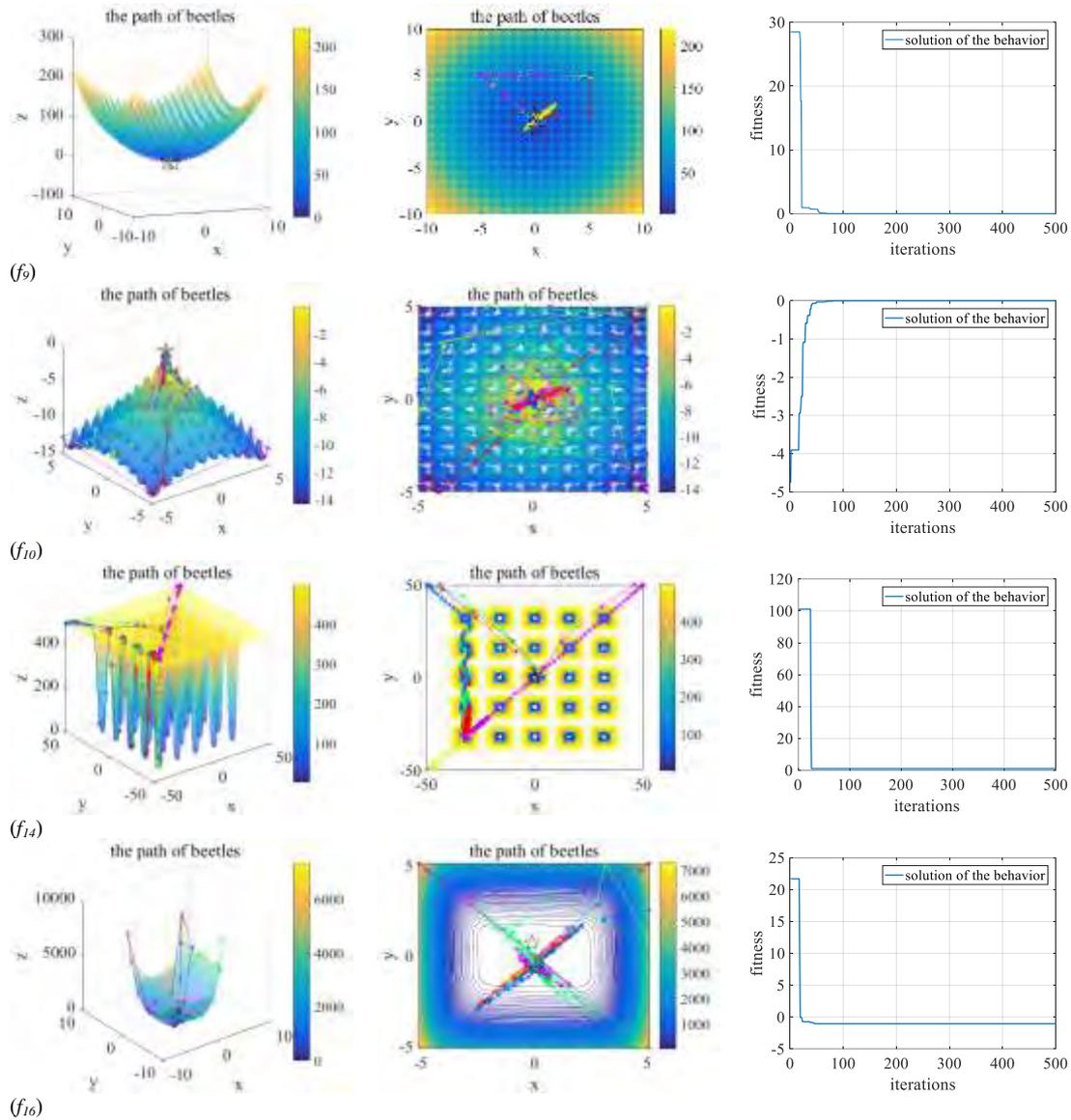

($f_9$)

($f_{10}$)

($f_{14}$)

($f_{16}$)

**Fig. 5** Behavior of BSO on the 2D benchmark problems

### 3.2 Quantitative Results and Discussion

The above discussion proves that the proposed algorithm can solve the optimization problem, but pure qualitative test can not prove the superiority of the algorithm. This section raises the dimensions of test functions other than fixed dimensions to 5 dimensions and gives quantified results. Table 5 gives the experimental results of the test function.

As shown in Table 5, when dealing with the unimodal functions, the processing speed of BSO is comparable to that of PSO, but it is obviously better than GA and GOA algorithm. In addition, compared with the other three algorithms, BSO algorithm is more stable in performance. Adding the beetle search mechanism in the process of optimization makes the algorithm have better global optimization performance, accelerates the convergence speed of the algorithm, and effectively avoids the phenomenon of "premature".

When dealing with multimode functions, BSO algorithm shows good performance again. Because multimodal functions have multiple local optimal solutions, the results can be directed to show that BSO algorithm is effective and efficient in avoiding local optimal solutions.

For the fixed-dimension multimodal functions, the proposed algorithm gives very competitive results. The

BSO algorithm has the ability to balance the exploration and exploitation of the individual and can solve more challenging problems.

**Table 5** Comparision of optimization results obtained for the unimodal, multimodal, and fixed-dimension multimodal functions

| F | BSO | | | PSO | | | GA | | | GOA | | |
|---|---|---|---|---|---|---|---|---|---|---|---|---|
| | *ave* | *std* | *ave_time*(s) | *ave* | *std* | *ave_time*(s) | *ave* | *std* | *ave_time*(s) | *ave* | *std* | *ave_time*(s) |
| F1 | 0 | 9.36E-76 | 0.5153 | 0 | 0 | 0.4597 | 0.0025 | 0.0017 | 3.7335 | 0.4004 | 0.3342 | 144.5615 |
| F2 | 1.02E-04 | 3.92E-04 | 0.6127 | 1.3333 | 3.4575 | 0.5099 | 0.008 | 0.0068 | 3.7362 | 1.3612 | 2.0519 | 29.6388 |
| F3 | 0 | 3.31E-72 | 0.8765 | 1.67E+02 | 912.8709 | 0.636 | 7.66E+03 | 2.34E+03 | 6.0088 | 0 | 0 | 29.8757 |
| F4 | 3.55E-09 | 1.07E-08 | 0.4999 | 0 | 0 | 0.459 | 15.7727 | 4.8173 | 3.6927 | 2.50E-05 | 1.20E-05 | 29.5846 |
| F5 | 0.6578 | 1.4017 | 0.6432 | 1.51E+04 | 3.41E+04 | 0.5247 | 43.927954 | 32.6768 | 3.7723 | 3.01E+03 | 1.64E+04 | 29.5752 |
| F6 | 0 | 0 | 0.5081 | 0 | 0 | 0.4591 | 0.0007 | 0.0011 | 3.7253 | 0 | 0 | 29.5096 |
| F7 | 5.17E-04 | 4.47E-04 | 0.6382 | 2.98E-04 | 0.0003 | 0.5219 | 0.0019 | 0.0009 | 3.9028 | 0.0737 | 0.1023 | 29.5672 |
| F8 | -1.79E+03 | 173.3453 | 0.6503 | -1.40E+03 | 85.7482 | 0.532 | -9.78E+03 | 373.5056 | 3.8002 | -1.74E+03 | 183.2 | 29.7437 |
| F9 | 0.4311 | 0.9305 | 0.5215 | 5.1785 | 9.0057 | 0.4683 | 59.7404 | 8.75764 | 3.7708 | 5.3052 | 2.9227 | 29.5213 |
| F10 | 0.1097 | 0.4177 | 0.6282 | 4.6379 | 8.4257 | 0.5253 | 0.007 | 0.0051 | 3.7441 | 0.6931 | 0.9474 | 29.5833 |
| F11 | 0.1267 | 0.0849 | 0.7203 | 0.1348 | 0.0926 | 0.5779 | 0.0725 | 0.1001 | 3.7637 | 0.1227 | 0.0638 | 29.7993 |
| F12 | 7.00E-06 | 3.76E-05 | 1.424 | 0 | 0 | 0.9052 | 36.1241 | 9.0446 | 4.0032 | 0.0011 | 0.0059 | 29.9328 |
| F13 | 0.0011 | 0.0034 | 1.4382 | 0 | 0 | 0.9123 | 57.65 | 12.9744 | 4.0068 | 0.0022 | 0.0044 | 29.9379 |
| F14 | 0.998 | 1.54E-16 | 1.9211 | 0.998 | 0 | 3.1104 | 0.998 | 0 | 3.8205 | 0.998 | 0 | 12.3002 |
| F15 | 0.0015 | 0.0036 | 0.586 | 0.0042 | 0.0117 | 0.4993 | 0.0039 | 0.00718 | 1.5158 | 0.0035 | 0.0067 | 19.9701 |
| F16 | -1.0316 | 6.71E-16 | 0.4534 | -1.0316 | 0 | 0.4272 | -1.0316 | 0 | 1.2441 | -1.0316 | 0 | 10.306 |
| F17 | 0.3979 | 0 | 0.5045 | 0.3979 | 0 | 0.5767 | 0.3979 | 0 | 1.2171 | 0.3979 | 0 | 10.2556 |
| F18 | 3 | 1.03E-15 | 0.3853 | 3 | 0 | 0.4031 | 3.9 | 4.9295 | 1.2144 | 5.7 | 14.7885 | 10.3014 |
| F19 | -3.8609 | 0.0034 | 0.8683 | -3.6913 | 0.1247 | 0.64 | -3.8627 | 0 | 1.5927 | -3.8369 | 0.1411 | 20.205 |
| F20 | -3.1256 | 0.3735 | 0.8685 | -2.1198 | 0.5567 | 0.6541 | -3.2625 | 0.0605 | 1.894 | -3.2698 | 0.0607 | 29.4666 |
| F21 | -9.8164 | 1.2818 | 0.5519 | -1.0902 | 0.8326 | 0.9072 | -5.9724 | 3.37309 | 1.9346 | -7.0499 | 3.2728 | 20.2475 |
| F22 | -10.0513 | 1.3381 | 0.6845 | -1.0196 | 0.4063 | 1.0713 | -7.3119 | 3.4237 | 2.1298 | -7.3062 | 3.4705 | 20.4859 |
| F23 | -9.1069 | 2.4111 | 0.9185 | -1.2161 | 0.6276 | 1.3545 | -5.7112 | 3.5424 | 2.4214 | -8.6298 | 3.0277 | 20.5744 |

## 3.3 Analysis of Convergence Behavior

Convergence curves of BSO,GA,GOA and PSO are compared in Fig.6 for all of the test functions. The figure shows that BSO has good processing ability for unimodal functions, multimodal functions and fixed-dimension functions, and the processing process is very stable. Especially when solving more complex fixed-dimension functions, BSO shows more obvious advantage than other algorithms. It can be seen that BSO is enough competitive with other state-of-the-art meta-heuristic algorithms.

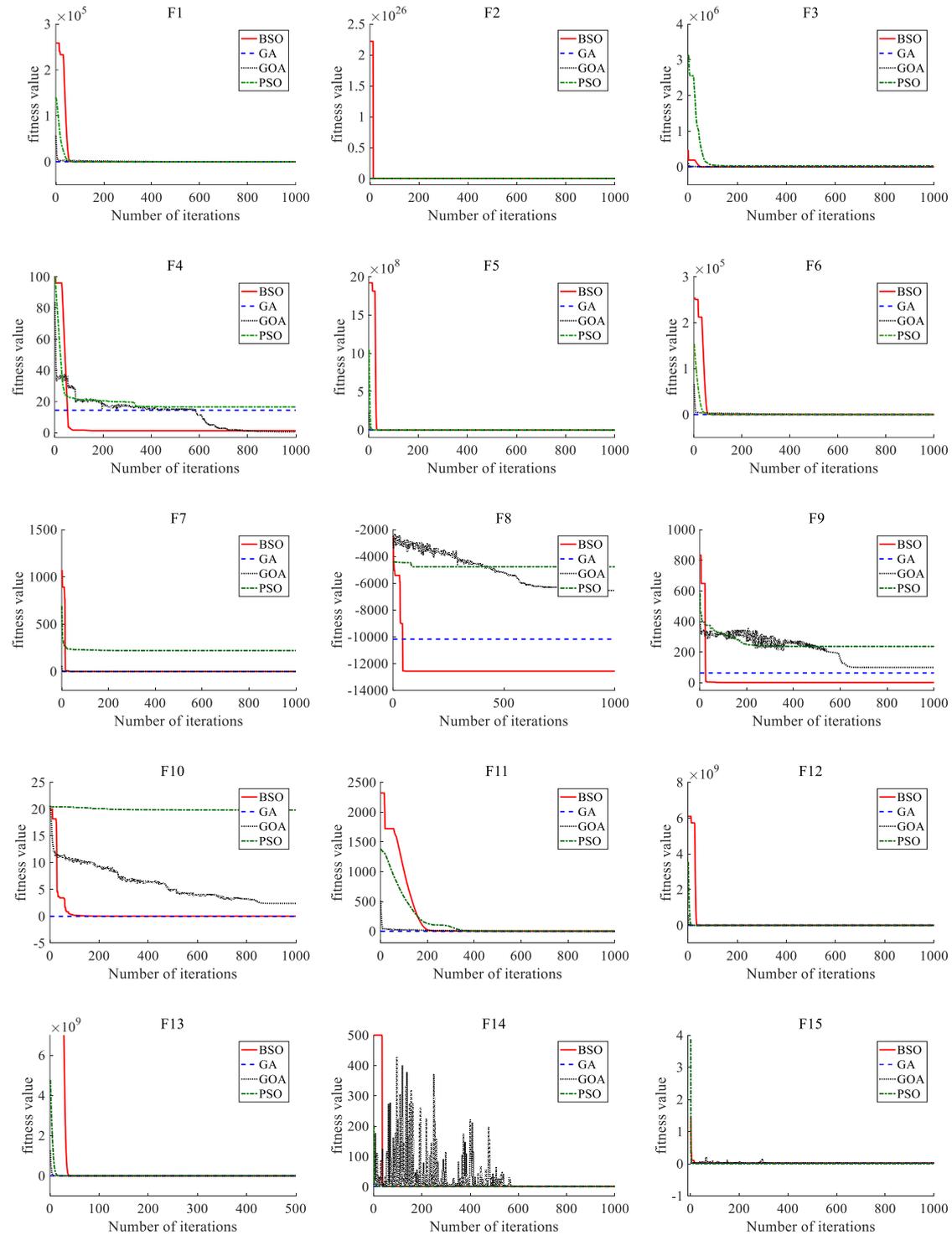

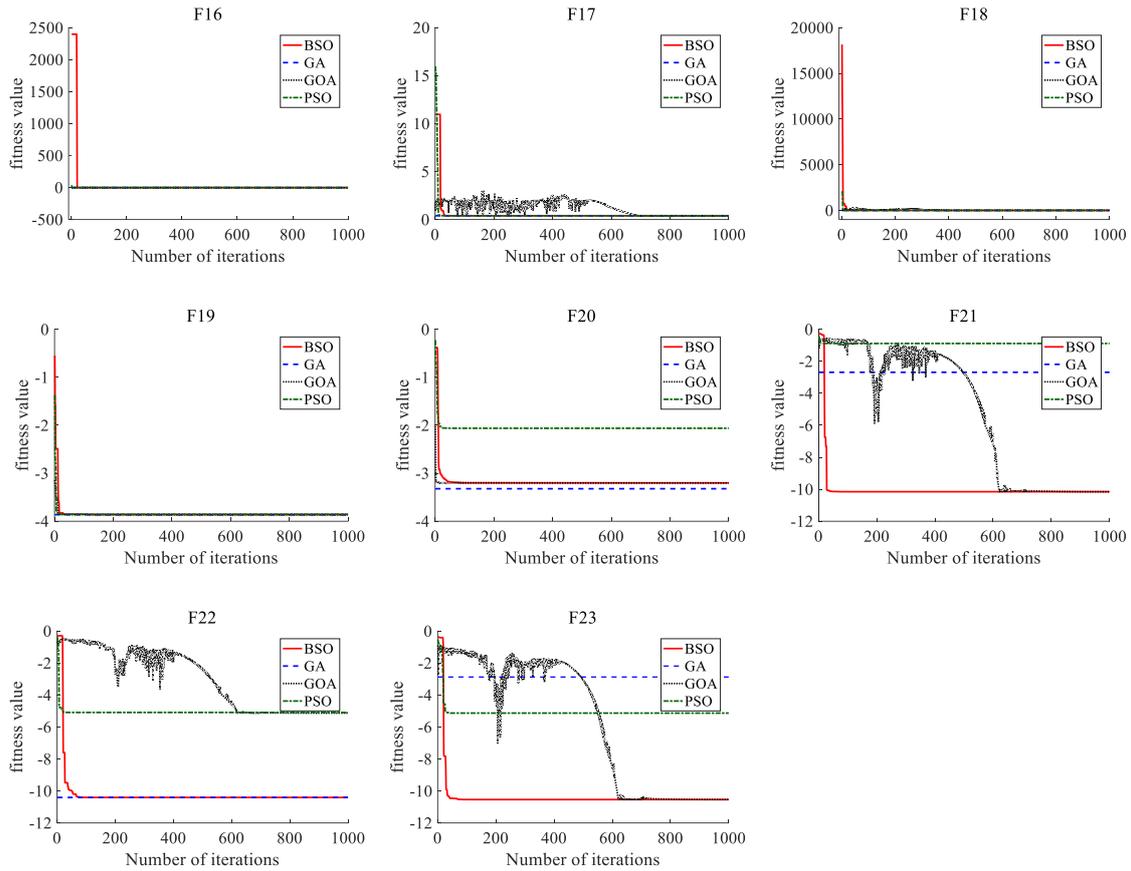

**Fig. 6** Comparison of convergence curves of BSO and literature algorithms obtained in all of the benchmark problems

As a summary, the results of this section revealed different characteristics of the proposed BSO algorithm. Efficient and stable search capabilities benefit from beetle-specific optimization features. The increase in the exploration function of the left and right must greatly improve the stability of the search, making the exploration and exploitation capabilities more balanced, and the BSO can handle better for high-dimensional and more complex problems. Overall, the success rate of the BSO algorithm seems to be higher in solving challenging problems. In the next sections, BSO performance is validated on more challenging multi-objective issues.

## 4 BSO for Multi-objective Optimization

In order to better illustrate the superiority and competitiveness of BSO algorithm in solving constrained optimization problems, two multi-objective functions in BAS algorithm are used in this paper, and the results are compared with the results of other algorithms.

### 4.1 BSO for a Pressure Vessel Design Problem

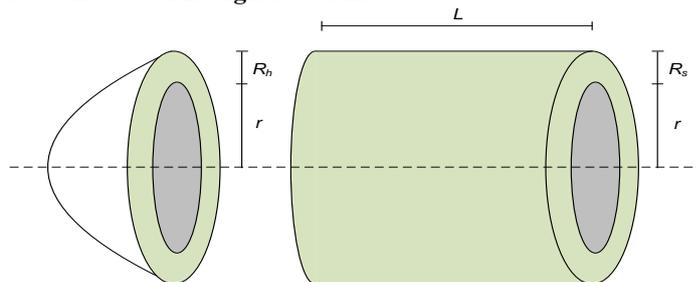

**Fig. 7** Schematic of pressure vessel

As shown in Fig.7, two hemispheres cover the ends of the cylinder to form a pressure vessel. The goal is to minimize the total cost including material costs, welding costs and molding costs[60]:

$$\text{minimize} \quad f_{\cos t}(x) = 0.6224 x_1 x_3 x_4 + 1.7781 x_2 x_3^2 + 3.1661 x_1^2 x_4 + 19.84 x_1^2 x_3$$

There are four variables in pressure vessel problem where $x_1$ is the thickness of the shell($R_s$), $x_2$ is the thickness of the head($R_h$), $x_3$ is the inner radius ($r$), and $x_4$ is the length of the section of the cylinder of the container ($L$). $R_s$ and $R_h$ are integral times of 0.0625, the available thickness of rolled steel plates, and $r$ and $L$ are continuous.

The constraint function can be stated as follows:

$$\begin{aligned}
s.t. \quad & g_1(x) = -x_1 + 0.0193 x_3 \leq 0, \\
& g_2(x) = -x_2 + 0.00954 x_3 \leq 0, \\
& g_3(x) = -\pi x_3^2 x_4 - \frac{4}{3}\pi x_3^3 + 1296000 \leq 0, \\
& g_4(x) = x_4 - 240 \leq 0, \\
& x_1 \in \{1,2,3,...,99\} \times 0.0625, \\
& x_2 \in \{1,2,3,...,99\} \times 0.0625, \\
& x_3 \in [10,200], \\
& x_4 \in [10,200].
\end{aligned}$$

Table 6 illustrates the best results obtained by the BSO algorithm just using 100 iterations and other various existing algorithm to solve the pressure vessel optimization. And most of these results are taken from Jiang et al.(2017).The results show that the best results of BSO algorithm are better than most existing algorithms and in the case where the population number is properly selected (we suggest 50 individuals), the convergence rate is faster and has good The robustness. The BSO algorithm iterative process is shown in Fig. 8.

**Table 6** comparisons results for pressure vessel function

| methods | x1 | x2 | x3 | x4 | g1(x) | g2(x) | g3(x) | g4(x) | f* |
|---|---|---|---|---|---|---|---|---|---|
| [61] | 0.8125 | 0.4375 | 42.0984 | 176.6378 | -8.8000e-7 | -0.0359 | -3.5586 | -63.3622 | 6059.7258 |
| [62] | 1.0000 | 0.6250 | 51.2519 | 90.9913 | -1.0110 | -0.1360 | -18759.75 | -149.009 | 7172.3000 |
| [63] | 0.8125 | 0.4375 | 42.0870 | 176.7791 | -2.210e-4 | -0.0360 | -3.5108 | -63.2208 | 6061.1229 |
| [64] | 1.0000 | 0.6250 | 51.0000 | 91.0000 | -0.0157 | -0.1385 | -3233.916 | -149.0000 | 7079.0370 |
| [65] | 0.8125 | 0.4375 | 41.9768 | 182.2845 | -0.0023 | -0.0370 | -22888.07 | -57.7155 | 6171.0000 |
| [66] | 0.9375 | 0.5000 | 48.3290 | 112.6790 | -0.0048 | -0.0389 | -3652.877 | -127.3210 | 6410.3811 |
| [67] | 0.8125 | 0.4375 | 40.3239 | 200.0000 | -0.0343 | -0.05285 | -27.10585 | -40.0000 | 6288.7445 |
| [68] | 1.1250 | 0.6250 | 58.1978 | 44.2930 | -0.0018 | -0.0698 | -974.3 | -195.707 | 7207.4940 |
| [69] | 1.1250 | 0.6250 | 48.9700 | 106.7200 | -0.1799 | -0.1578 | 97.760 | -132.28 | 7980.8940 |
| [70] | 1.1250 | 0.6250 | 58.2789 | 43.7549 | -0.0002 | -0.0690 | -3.71629 | -196.245 | 7198.4330 |
| [44] | 0.8125 | 0.4375 | 42.0936 | 176.7715 | -9.43e-05 | -0.0359 | -413.6252 | -63.2285 | 6062.0468 |
| BSO | 0.8125 | 0.4375 | 42.0984 | 176.6366 | 0.0000 | -0.0359 | 0.0000 | -63.3634 | 6059.7000 |

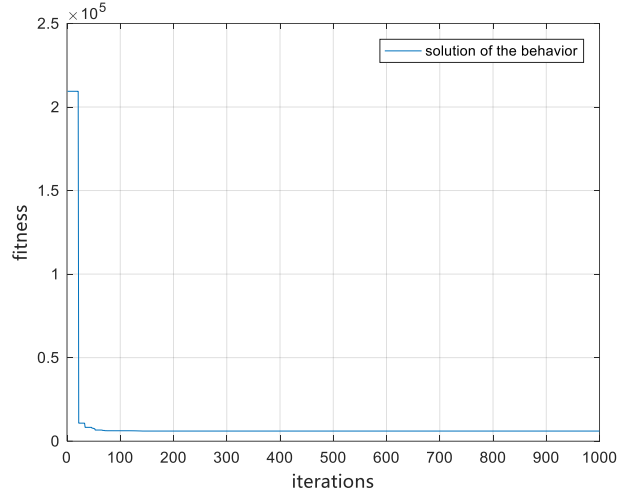

**Fig. 8** Iteration process for pressure vessel design problem

### 4.2 BSO for Himmelblau's Optimization Problem

This problem is proposed by Himmelblau[71] and is a common function for nonlinear constrained optimization problems. It is widely used in the optimization field. It consists of five variables, three equality constraints and six inequality constraints. The specific forms are as follows:

minimize $f(x) = 5.3578547 x_3^2 + 0.8356891 x_1 x_5 + 37.29329 x_1 - 40792.141,$

s.t. $g_1(x) = 85.334407 + 0.0056858 x_2 x_5 + 0.00026 x_1 x_4 - 0.0022053 x_3 x_5,$

$g_2(x) = 80.51249 + 0.0071317 x_2 x_5 + 0.0029955 x_1 x_2 + 0.0021813 x_3^2,$

$g_3(x) = 9.300961 + 0.0047026 x_3 x_5 + 0.0012547 x_1 x_3 + 0.0019085 x_3 x_4,$

$0 \le g_1(x) \le 92,$

$90 \le g_2(x) \le 110,$

$20 \le g_3(x) \le 25,$

$78 \le x_1 \le 102,$

$33 \le x_2 \le 45,$

$27 \le x_3 \le 45,$

$27 \le x_4 \le 45,$

$27 \le x_5 \le 45.$

Table 7 shows the performance results of the existing algorithm and the BSO algorithm. The number of iterations is set to 100. Evidently, the best result generated from the BSO shows the most excellent performance among all the results listed in Table . The above experiments justify that the proposed BSO algorithm is effective to handle constraint optimum problem and could achieve a good performance with high convergence rate. In the experiment process, when the population size is 50 and the number of iterations is 1000, the effect is the most stable. The BSO algorithm iterative process is shown in Figure 9.

**Table 7** comparisons results for himmelblau function

| methods | $x_1$ | $x_2$ | $x_3$ | $x_4$ | $x_5$ | $g_1(x)$ | $g_2(x)$ | $g_3(x)$ | $f_*$ |
|---|---|---|---|---|---|---|---|---|---|
| [72] | 78.00 | 33.00 | 29.995256 | 45.00 | 36.775813 | 92.00 | 98.8405 | 20.0000 | -30665.54 |
| [73] | 78.00 | 33.00 | 29.995256 | 45.00 | 36.775813 | 92.00 | 98.8405 | 20.0000 | -30665.539 |
| [74] | 81.49 | 34.09 | 31.2400 | 42.20 | 34.3700 | 91.78 | 99.3188 | 20.0604 | -30183.576 |
| [75] | 78.00 | 33.00 | 29.995256 | 45.00 | 36.7258 | 90.71 | 98.8287 | 19.9929 | -30665.539 |
| [44] | 78.00 | 33.00 | 27.1131 | 45.00 | 45.0000 | 92.00 | 100.4170 | 20.0206 | -31011.3244 |
| BSO | 78.00 | 33.00 | 27.0710 | 45.00 | 44.9692 | 92.00 | 100.4048 | 20.0000 | -31025.5563 |

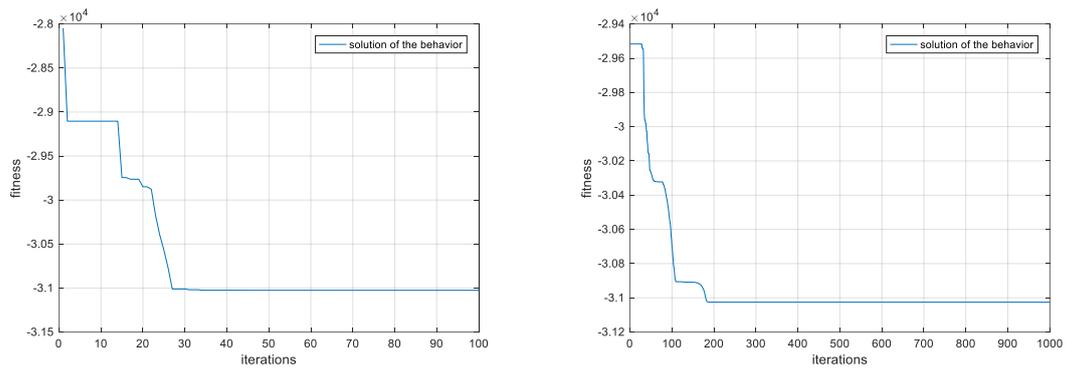

**Fig. 9** Iteration process for himmelblau's optimization problem

## 5 Conclusions

This paper proposes a new meta-heuristic algorithm called beetle group optimization. The algorithm combines the beetle's foraging mechanism with the group optimization algorithm, and establishes a mathematical model and applies it to unimodal functions, multimodal functions, fixed-dimension multimodal benchmark functions. The results show that compared with the current popular optimization algorithms, the BSO algorithm can still give very competitive results, and has good robustness and running speed. In addition, the BSO algorithm also exhibits higher performance when dealing with nonlinear constraints. Compared with other optimization algorithms, BSO can handle multi-objective optimization problems efficiently and stably.

Finally, in the research process, we found that the change in step size and speed will affect the efficiency and effectiveness of BSO optimization. Therefore, in the next work, we will further study the impact of different parameter settings on BSO.